\documentclass[10pt,twocolumn,letterpaper]{article}

    \usepackage[pagenumbers]{cvpr} % To force page numbers, e.g. for an arXiv version

\usepackage{graphicx}
\usepackage{amsmath}
\usepackage{amssymb}
\usepackage{booktabs}
\usepackage{makecell,multirow}

\usepackage[pagebackref,breaklinks,colorlinks]{hyperref}

\usepackage[capitalize]{cleveref}
\crefname{section}{Sec.}{Secs.}
\Crefname{section}{Section}{Sections}
\Crefname{table}{Table}{Tables}
\crefname{table}{Tab.}{Tabs.}

\begin{document}

\title{Traceable and Authenticable Image Tagging for Fake News Detection}

\author{Ruohan Meng{$^{1,3,4}$},~ Zhili Zhou{$^{2}$\thanks{Corresponding Author}},~ Qi Cui{$^{1,5}$},~ Kwok-Yan Lam{$^{3}$},~ Alex Kot{$^{4}$}\\\vspace{-8pt}{\small~}\\
{$^{1}$}School of Computer and Software, Nanjing University of Information Science and Technology, China\\ {$^{2}$}Institute of Artificial Intelligence and Blockchain, Guangzhou University, China\\ {$^{3}$}Computer Science and Engineering, Nanyang Technological University, Singapore\\
{$^{4}$}Rapid-Rich Object Search (ROSE) Lab, Nanyang Technological University, Singapore\\
{$^{5}$}Centre for Computer Vision and Deep Learning, University of Windsor, Canada\\
{\small{ \tt{ruohanmeng.melody@gmail.com}}},
{\small{ \tt{zhou\_zhili@163.com}}},
{\small{ \tt{cuiqiloveslife@gmail.com}}},\\
{\small{ \tt{kwokyan.lam@ntu.edu.sg}}},
{\small{ \tt{eackot@ntu.edu.sg}}}
}

\iffalse

\fi

\maketitle

%%%%%%%%% ABSTRACT
\begin{abstract}
  To prevent fake news images from misleading the public, it is desirable not only to verify the authenticity of news images but also to trace the source of fake news, so as to provide a complete forensic chain for reliable fake news detection. To simultaneously achieve the goals of authenticity verification and source tracing, we propose a traceable and authenticable image tagging approach that is based on a design of Decoupled Invertible Neural Network (DINN). The designed DINN can simultaneously embed the dual-tags, \textit{i.e.}, authenticable tag and traceable tag, into each news image before publishing, and then separately extract them for authenticity verification and source tracing. Moreover, to improve the accuracy of dual-tags extraction, we design a parallel Feature Aware Projection Model (FAPM) to help the DINN preserve essential tag information. In addition, we define a Distance Metric-Guided Module (DMGM) that learns asymmetric one-class representations to enable the dual-tags to achieve different robustness performances under malicious manipulations. Extensive experiments, on diverse datasets and unseen manipulations, demonstrate that the proposed tagging approach achieves excellent performance in the aspects of both authenticity verification and source tracing for reliable fake news detection and outperforms the prior works. 
\end{abstract}

%%%%%%%%% BODY TEXT
\section{Introduction}
\label{sec:intro}
\begin{figure}
	\centering
	\includegraphics[width=0.95\linewidth]{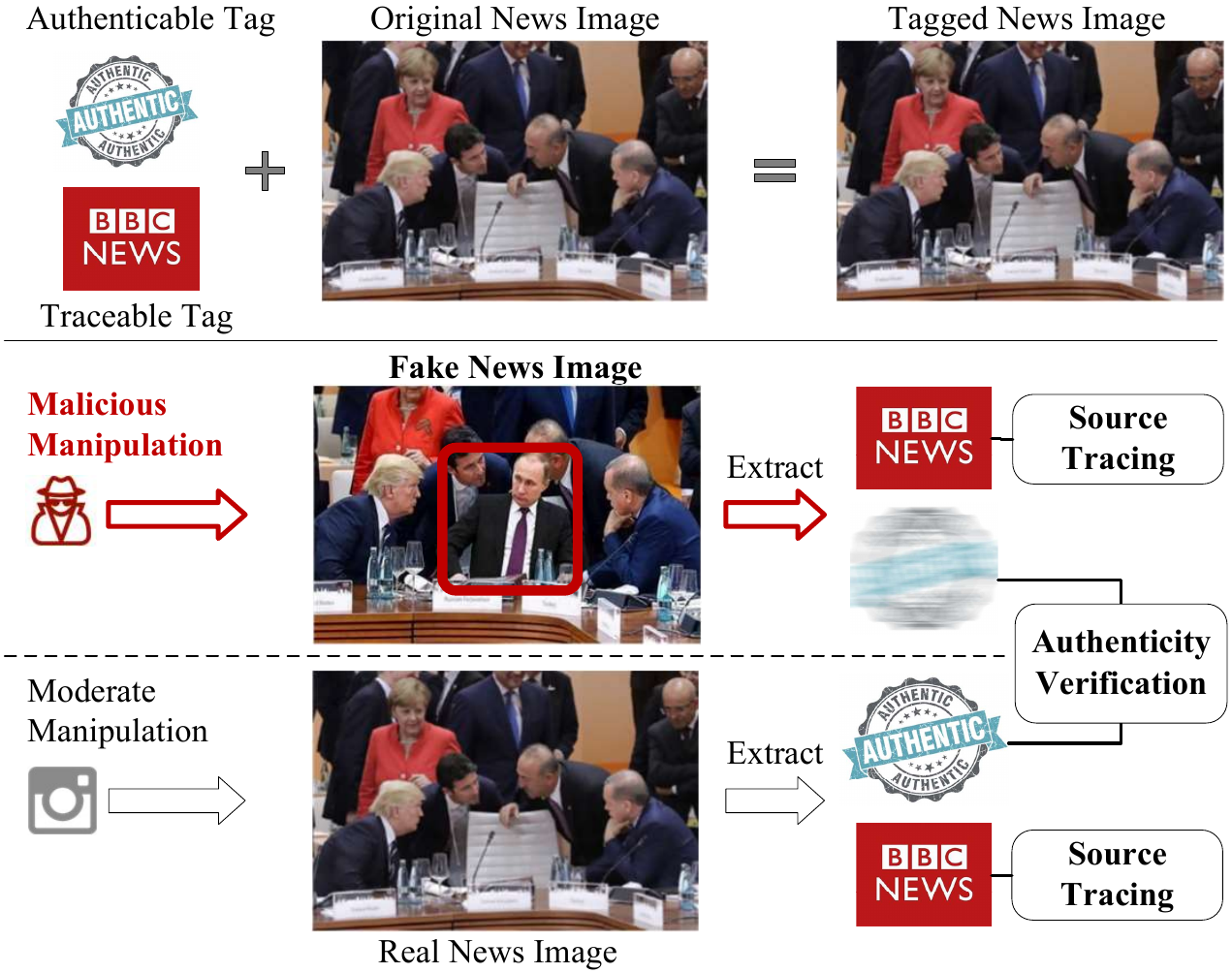}
	\vspace{-0.2cm}
	\caption{The framework of image tagging for reliable fake news detection. Authenticable tags and traceable tags are simultaneously embedded into news images to generate tagged images. The dual-tags should be correctly extracted from real news images. However, as for fake news images, authenticable tags should be fragile to malicious manipulations by extracting wrong tags to verify the news authenticity, and traceable tags should be robust to malicious manipulations to achieve source tracing.}
	\label{fig:diagram}
	\vspace{-0.3cm}
\end{figure}
In the We-media era, news content with malicious manipulation, \textit{i.e.}, fake news, is easily produced and distributed on social media. A large amount of fake news, especially provocative fake news images, is undermining public credibility and influencing social stability, resulting in many serious social security issues~\cite{13}. To detect fake news images, the existing methods usually design detectors by capturing the traces of malicious manipulations such as textual information~\cite{6,10,11}, visual features~\cite{32,33,35}, and multi-modal fusion of features~\cite{14,18} from news images. Although those methods have achieved desirable performance for verifying the authenticity of news images, they do not take source tracing into consideration. The news source traceability is essential for the public to know where the news is published to enhance the trustworthiness in news. Thus, the authenticity verification and source traceability can jointly provide a complete forensic chain for reliable fake news detection.

In the existing deep watermarking methods, robust watermarking~\cite{37,70,30} or fragile watermarking~\cite{31,50,68} can either achieve copyright traceability or content integrity authentication. To achieve the two purposes simultaneously, some dual-watermarking methods~\cite{71,73,74,82} have been proposed to manually embed the two types of watermarks, and thus they suffer from two common issues. 1) Once the two types of watermarks are manually embedded into an image, they would affect each other, which makes them hard to decouple and extract; 2) It is hard to guarantee the two types of embedded watermarks achieve different robustness performances to malicious manipulations for different purposes.

Therefore, it is an interesting and challenging task to achieve the goals of news authenticity verification and source tracing simultaneously. To this end, we propose an image tagging approach based on a design of Decoupled Invertible Neural Network (DINN). In this approach, the dual-tags, \textit{i.e.}, traceable tag and authenticable tag, can be invisibly embedded into the news images before publishing, and be separately extracted for authenticity verification and source tracing, respectively. The framework of the proposed approach is shown in \cref{fig:diagram}. In summary, the contributions of this paper include: 

\begin{itemize}
	\item We propose a novel proactive image tagging scheme to achieve both news image authenticity verification and source tracing for reliable fake news detection.
	\item We design a Decoupled Invertible Neural Network (DINN) to simultaneously embed two types of invisible tags, \textit{i.e.}, authenticable tag and traceable tag, into each news image, and extract the two tags from all manipulated news images separately without interfering with each other.
	\item We design a double Feature Aware Projection Model (FAPM) and a Distance Metric-Guided Module (DMGM) to improve the recovery accuracy of embedded tags and enable the dual-tags to achieve different robustness performances to malicious manipulations1. 
\end{itemize}

%-------------------------------------------------------------------------
%------------------------------------------------------------------------
\section{Related Works}
\label{sec:formatting}

%-------------------------------------------------------------------------
\begin{figure*}
	\centering
	\includegraphics[width=\linewidth]{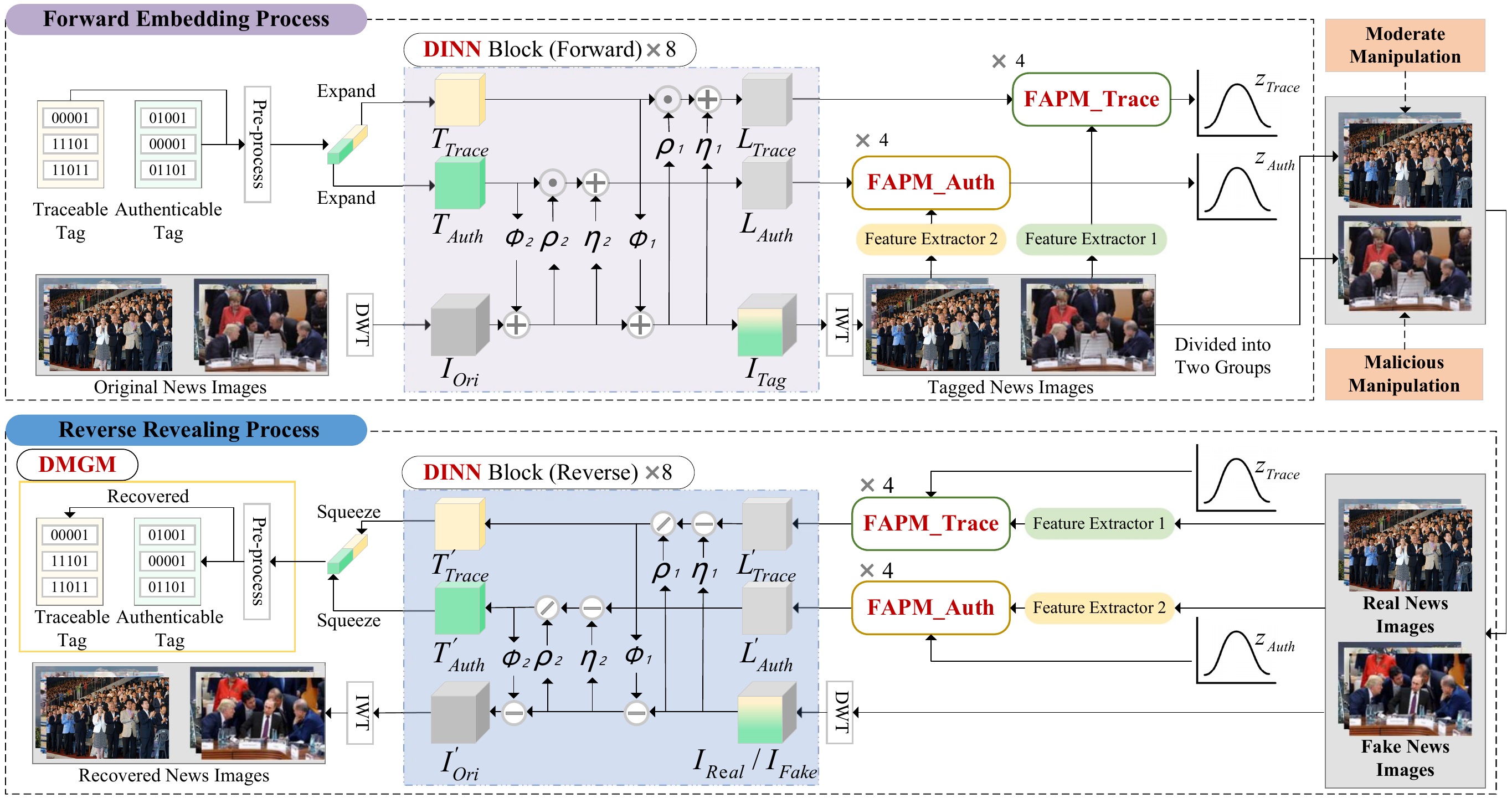}
	
	\caption{Overview of the traceable and authenticable image tagging approach based on the designed DINN for reliable fake news detection in the training process. In the inference process, the embedding and revealing processes are the same as the training process.}
	\vspace{-0.4cm}
	\label{fig:Overview}
\end{figure*}

\iffalse
\begin{table}
	\centering
	\caption{Summary of notations in this paper.}
	\begin{tabular}{@{}lc@{}}
		\toprule
		Notation & Description \\
		\midrule
		$I_{\textit {ori }}$ & Preprocessed original news images \\
		
		$T_{\textit {Trace }}$ and $T_{\textit {Auth }}$ &
		\makecell[c]{Preprocessed traceable tag and\\ authenticable tag}\\
		
		$I_{\textit {Tag }}$ & The tagged news images\\
		
		$L_{\textit {Trace }}$ and $L_{\textit {Auth }}$ & \makecell[c]{Lost information of $T_{\textit {Trace }}$ \\and $T_{\textit {Auth }}$ in the embedding process}\\
		
		$L_{\textit {Trace }}^{\prime}$ and $L_{\textit {Auth }}^{\prime}$ & \makecell[c]{Revealed lost information of $T_{\textit {Trace }}$ and \\$T_{\textit {Auth }}$ from pre-defined distribution}\\
		
		$I_{\textit {Real }}$ & \makecell[c]{Real news images under\\ benign maniplation}\\
		
		$I_{\textit {Fake }}$ & \makecell[c]{Fake news images under\\ malicious maniplation}\\
		
		$T_{\textit {Trace }}^{\textit {ext r }}$ and $T_{\textit {Trace }}^{\textit {ext f }}$ & \makecell[c]{Extracted traceable tags \\ from real news images and fake news\\ images, respectively}\\
		
	    $T_{\textit {Auth }}^{\textit {ext f }}$ and $T_{\textit {Auth }}^{\textit {ext f }}$& \makecell[c]{Extracted authenticable tags\\ from real news images and fake news\\ images, respectively}\\
		\bottomrule
	\end{tabular}
	
	\label{tab:notations}
\end{table}
\fi
\subsection{Fake News Detection}

Generally, existing fake news detection methods try to capture unusual patterns by extracting text information~\cite{4,6,10,11}, visual features~\cite{13,32,33,35}, and multi-modal fusion of features ~\cite{12,14,15,18} from news content. As the one main component of news, textual content~\cite{4,11} is widely used as the main domain for detecting fake news. Ma \textit{et al.}~\cite{4} adopted a 2-layer Gated Recurrent Unit model to extract the hidden representations, which can capture the variation of contextual information of relevant posts over a period. Besides, there are a lot of algorithms that utilize different NLP models~\cite{5,8}, autoencoder~\cite{7}, and GAN~\cite{9} to learn more comprehensive representations of textual content to detect fake news. Moreover, several algorithms take stance classification~\cite{6,11} and writing style~\cite{10} into account to detect fake news. 

In addition, as another component of news, visual features such as news images or videos are more easily disseminated and maliciously manipulated. Some methods~\cite{32,33,34} analyze the basic statistical features from news image amount, image type, and image popularity, while some other methods extract forensic features for fake image detection~\cite{35,36}. Besides, Qi \textit{et al.}~\cite{13} proposed Multi-domain Visual Neural Network (MVNN) to capture the complex patterns of fake news images in the frequency domain and extract visual features in the pixel domain, which improves the performance of fake news image detection. Since the representation capability of one specific modal feature is limited, some multi-modal methods~\cite{12,14,15,18} combine the visual features and textual features to explore their consistency to verify the fake news. 

All the above content-based fake news detection methods detect fake news passively by analyzing textural features and/or visual features of news. Although these methods improve fake news detection from different levels, these passive detectors do not consider the news traceability to achieve reliable fake news detection. Hence, to provide a reliable forensic chain for fake news detection, we propose a proactive tagging scheme by embedding traceable and authenticable tags into original news images to detect fake images and trace the news source.

%-------------------------------------------------------------------------
\subsection{Watermarks}

Digital image watermarking can achieve traceability and content integrity authentication. To achieve source traceability, robust watermarking~\cite{30,37,69,70} is proposed to resist all kinds of attacks. Zhu \textit{et al.}~\cite{37} proposed the end-to-end trainable framework based on autoencoder, \textit{i.e.}, HiDDeN, which inserts noise layers between encoding and decoding to simulate image distortion. Subsequently, Jia \textit{et al.}~\cite{70} proposed Mini-Batch of Real and Simulated JPEG compression (MBRS) to enhance the JPEG robustness based on autoencoder. However, the two sub-networks in autoencoder-based methods have two sets of parameters, which cause color distortion and low invisibility. HiNet~\cite{54} and ISN~\cite{55} have been proposed to integrate the embedding and extraction process into an invertible model to improve image quality and security, which are based on INN. Besides, RIIS~\cite{30} introduced a conditional normalizing flow to model the distribution of high-frequency components as lost information and add a distortion model to simulate the distortion for robust watermarking. 

Besides, to achieve content integrity authentication, fragile watermarks-based methods~\cite{31,50,68} are proposed. Neekhara \textit{et al.}~\cite{31} proposed to proactive embed a semi-fragile neural watermark into real images for digital media authentication. Besides, Asnani \textit{et al.}~\cite{50} proposed a template estimation scheme to proactively detect manipulated images by embedding orthogonal and learnable templates into real images. It achieves better image manipulation detection performance on unseen generative models.

Some dual-watermarking methods~\cite{71,52,73,74,82} have been proposed to embed different watermarks to achieve multi-purposes of image protection. Lu and Liao~\cite{74} embedded the robust and fragile watermarks in an image for copyright protection and image authentication. For the same purposes, Liu \textit{et al.}~\cite{73} embedded one watermark in YCbCr color space by using DWT and another watermark in RGB components. Besides, TRLH~\cite{82} is proposed to embed fragile and blind dual-watermarks for image tamper detection and self-recovery based on lifting wavelet and halftoning techniques. However, these methods embed watermarks by extracting features manually and thus cannot flexibly handle different attacks. 

For fake news detection, the existing methods cannot be directly used to achieve good performance on both news traceability and authentication simultaneously. Therefore, we propose to embed the dual-tags, \textit{i.e.}, traceable tags and authenticable tags, into each news image to achieve both authenticity identification and source tracing to build a complete forensic chain in fake news detection.

%-------------------------------------------------------------------------
\section{Proposed Tagging Approach}
In this section, we propose a novel proactive tagging scheme by embedding dual-tags into each news image. The dual embedded tags can achieve different robustness performances to malicious manipulations. Then, they can be dully decoupled and extracted to achieve news authentication and traceability simultaneously for reliable fake news detection. 

\subsection{Network Architecture}

The architecture of our proposed approach is shown in \cref{fig:Overview}. It contains three main structures: 1) The forward embedding process, 2) Simulate manipulations attack in practical, and 3) The reverse revealing process. Note that \cref{fig:Overview} shows the training process of our proposed scheme, and the inference process is the same as the training process.

First, traceable tags and authenticable tags are embedded into news images with three branches and then are extracted from the news images under a variety of manipulations. In the process, the lost information of dual-tags with the conditional features extracted from tagged images is encoded into two sets of latent variables under Feature Aware Projection Model (FAPM) to improve the extraction accuracy of the embedded tags.

With the consideration of news characteristics, we simulate the manipulations that news images may undergo, including moderate manipulation and malicious manipulation. The former will not affect the news authenticity, such as compression, adjusting contrast, blur, etc.  While the latter will tamper with the content of news images to mislead the public, including splicing, copy-move, etc. Therefore, the images under moderate and malicious manipulations are defined as real images and false images, respectively. 

In the revealing process, with the help of Distance Metric-Guided Module
(DMGM), the pre-embedded traceable tags and authenticable tags are extracted with different accuracy from the manipulated news images for authenticity verification and source tracing. 

To facilitate the embedding process, we normalize the tags, such as logos, news information, etc., as bitstreams. We adopt the normalization strategy similar to that in IWN~\cite{57}.

\begin{figure}
	\centering
	\includegraphics[width=1\linewidth]{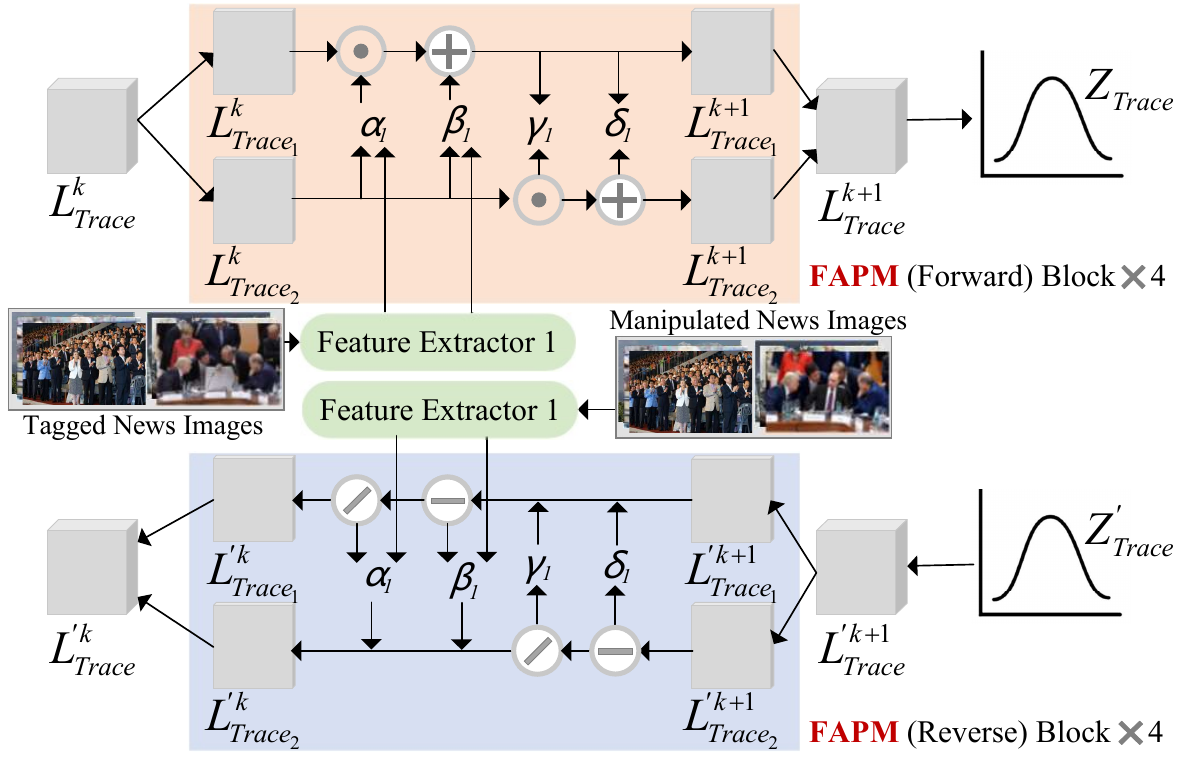}
	
	\caption{The architecture of the FAPM. It takes the traceable tag as an example to show the process.}
	\vspace{-0.5cm}
	\label{fig:fig3}
\end{figure}

%-------------------------------------------------------------------------
\subsection{Decoupled Invertible Neural Network (DINN)}

Different from the existing INN-based methods~\cite{54,55,30,57}, we design to embed the dual-tags, \textit{i.e.}, traceable tag and authenticable tag, and they should show different robustness performances under image manipulations. In our design, authenticable tags should be robust to moderate manipulations, while fragile to malicious manipulations. And traceable tags should be robust to all manipulations. 

The two branches of existing INNs share the same network parameters if using classical INN, and thus it is hard for those INNs to achieve different robustness performances for the two embedded tags.

To achieve our goals, we propose the Decoupled Invertible Neural Network (DINN) to decouple the embedding and revealing processes of traceable tag and authenticable tag, so that these dual-tags will not interfere each other. Before training the proposed network, each original news image is first decomposed by DWT with Haar wavelet kernel to obtain $I_{Ori}$, and traceable tag and authenticable tag are normalized to obtain $T_{Trace}$ and $T_{Auth}$. Then, $I_{Ori}$, $T_{Trace}$ and $T_{Auth}$ are served as the inputs of the forward embedding process to output the tagged news images $I_{Tag}$ and the corresponding lost information $L_{Trace}$ and $L_{Auth}$. Here, we encode $L_{Trace}$ and $L_{Auth}$ into a set of predefined distributed latent variables via conditional INNs for reserving more tag information. Meanwhile, the tagged news images would undergo moderate manipulations and malicious manipulations, resulting in real news images $I_{Real}$ and fake news images $I_{Fake}$, respectively. For the reverse revealing process of DINN, $I_{Real}$, $I_{Fake}$, and revealed lost information of $L_{Trace}$ and $L_{Auth}$ are used as inputs, and then the  $T_{Trace}$ and $T_{Auth}$ are extracted from both $I_{Real}$ and $I_{Fake}$ . According to the extraction accuracy rates of tags to verify the news source and authenticity.

As shown in \cref{fig:Overview}, DINN consists of several decoupled invertible blocks with the same structure. For the $n$-th decoupled invertible block, the input is $\left[T_{\text {Trace }}^n, T_{\text {Auth }}^n, I_{\text {Ori }}^n\right]$ and its output is $\left[T_{\text {Trace }}^{n+1}, T_{\text {Auth }}^{n+1}, I_{\text {Ori }}^{n+1}\right]$. Formally, the forward process is calculated as follows:
\begin{equation}
% \small
	I_{\textit {Ori }}^{n+1}=I_{\textit {Ori }}^n+\phi_1\left(T_{\textit {Trace }}^n\right)+\phi_2\left(T_{\textit {Auth }}^n\right)\;\;\;\;\;\;\;\;\;\;\;
	\label{eq:eq1}
\end{equation}
\begin{equation}
\begin{aligned}
	\;T_{\text {Auth }}^{n+1}&=T_{\text {Auth }}^n \odot \exp \left(\rho_2\left(I_{\text {Ori }}^n+\phi_2\left(T_{\text {Auth }}^n\right)\right)\right) \;\;\;\\
	&\;\;\;+\eta_2\left(I_{\text {Ori }}^n+\phi_2\left(T_{\text {Auth }}^n\right)\right)
\end{aligned}
%\begin{gathered}
%	T_{\text {Auth }}^{n+1}=T_{\text {Auth }}^n \odot \exp \left(\rho_2\left(I_{\text {Ori }}^n+\phi_2\left(T_{\text {Auth }}^n\right)\right)\right) \\
%	+\eta_2\left(\phi_2\left(T_{\text {Auth }}^n\right)\right)
%\end{gathered}
	\label{eq:eq2}
\end{equation}
\begin{equation}
	\;\;\;T_{\text {Trace }}^{n+1}=T_{\text {Trace }}^n \odot \exp \left(\rho_1\left(I_{\text {Ori }}^{n+1}\right)\right)+\eta_1\left(I_{\text {Ori }}^{n+1}\right)
	\label{eq:eq3}
\end{equation}
Where, $\phi_1(\cdot)$, $\phi_2(\cdot)$, $\rho_1(\cdot)$, $\rho_2(\cdot)$, $\eta_1(\cdot)$, and $\eta_2(\cdot)$ are sub-modules with convolution operations, $\exp (\cdot)$ is the Exponential function, and $\odot$ is the Hadamard product operation. 

In the reverse revealing process of DINN, the concatenated images of $I_{\textit {Real}}$ and $I_{\textit {Fake}}$, and revealed lost information is fed in, and then four tags including $T_{\textit {Trace}}$ and $T_{\textit {Auth}}$ from both  $I_{\textit {Real}}$ and $I_{\textit {Fake}}$ are extracted. Specifically, in the revealing process, the information inverse direction is from the $(n+1)$-th decoupled invertible block to $n$-th decoupled invertible block, as shown in \cref{fig:Overview}. For the $(n+1)$-th triplet invertible block, the inputs are the revealed lost information $L_{\textit {Trace }}^{n+1}$ and $L_{\textit {Auth }}^{n+1}$, which are randomly sampled from distributions.  $I_{\textit {Tag }}^{n+1}$ represents the manipulated news images, which is the concatenation of real news images $I_{\textit {Real}}$ and fake news images $I_{\textit {Fake}}$, as the input in the revealing process of DINN. Accordingly, the reverse process is calculated as follows:
\begin{equation}
	\begin{aligned}
	    T_{\textit {Trace }}^n&=\left(L_{\textit {Trace }}^{n+1}-\eta_1\left(I_{\textit {Tag }}^{n+1}\right)\right)\;\;\;\;\;\;\;\;\;\;\;\; \\
	    &\;\;\;\;\odot \exp \left(-  \rho_1\left(I_{\textit {Tag }}^{n+1}\right)\right)
	\label{eq:eq4}
	\end{aligned}
\end{equation}

\begin{equation}
	\begin{aligned}
		\;\;\;T_{\text {Auth }}^n&=\left(L_{\textit {Auth }}^{n+1}-\eta_2\left(I_{\textit {Tag }}^{n+1}-\phi_1\left(T_{\textit {Trace }}^n\right)\right)\right) \\
		&\;\;\;\;\;\odot \exp \left(-\rho_2\left(I_{\textit {Tag }}^{n+1}-\phi_1\left(T_{\textit {Trace }}^n\right)\right)\right)
	\end{aligned}
	\label{eq:eq5}
\end{equation}

\begin{equation}
	\begin{gathered}
		I_{\textit {Tag }}^n=I_{\textit {Tag }}^{n+1}-\phi_1\left(T_{\textit {Trace }}^n\right)-\phi_2\left(T_{\textit {Auth }}^n\right)
	\end{gathered}
	\label{eq:eq6}
\end{equation}

After the last revealing block, both the embedded traceable tag and authenticable tag are extracted from each news image. It is noted that the extracted authenticable tag from fake news images should be quite different from the original one. And other tags should be extracted correctly from other news images.

%-------------------------------------------------------------------------
\subsection{Feature Aware Projection Model (FAPM)}
Although the proposed DINN can decouple the embedding and extarcting of dual-tags, the desirable extraction accuracy performance still can not be guaranteed. To improve method usability, we design FAPM as the auxiliary to further support the decoupling process and guarantee extraction performance.

Inspired by the invertible image decolorization~\cite{58} and the model of content-aware noise projection in RIIS~\cite{30}, we encode some lost information of $L_{\textit {Trace}}$ and $L_{\textit {Auth}}$ into two sets of pre-defined distributed latent variables via conditional INNs to improve the extraction performance. The embedded $T_{\textit {Trace}}$ and $T_{\textit {Auth}}$ can be efficiently revealed by randomly re-sampling new sets of pre-defined distributed variables. 
\begin{figure}
	\centering
	\includegraphics[width=1\linewidth]{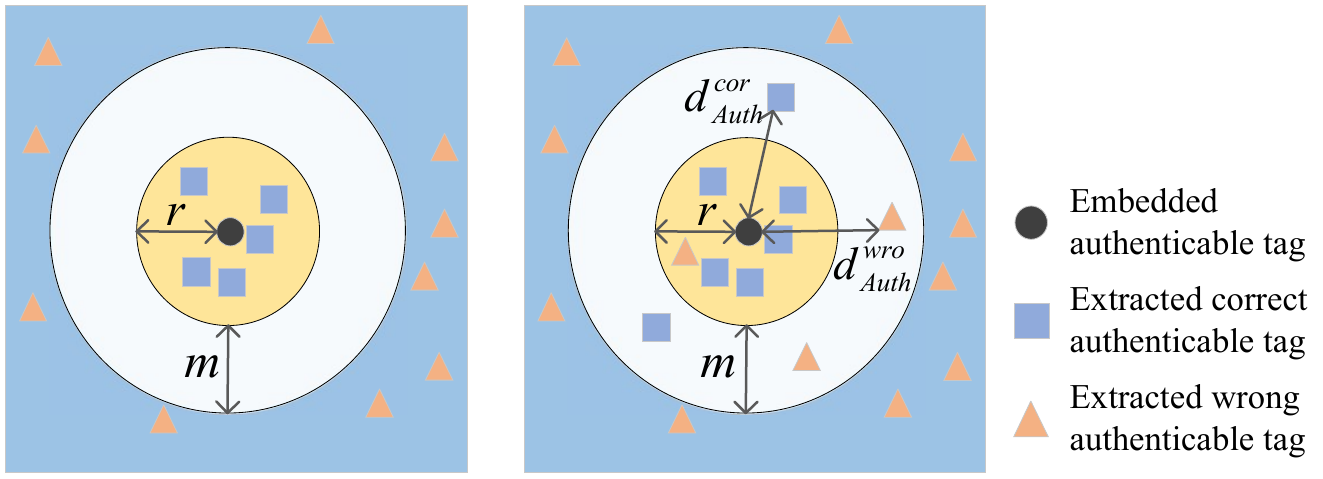}
	
	\caption{The architecture of the DMGM.}
	\label{fig:fig4}
	\vspace{-0.3cm}
\end{figure}

In the embedding process, we expect to preserve more tag information in lost information $L_{\textit {Trace}}$ and $L_{\textit {Auth}}$, to perfectly reveal the embedded tags. Besides, to completely decouple the dual-tags and achieve different extraction goals, we employ conditional INN to encode the lost information of $L_{\textit {Trace}}$ and $L_{\textit {Auth}}$. With the conditional features extracted from tagged images $I_{Tag}$, the lost information is converted to two sets of latent variables $z_{Trace}$ and $z_{Auth}$, which follow a pre-defined uniform distribution  $U(-1,1)$. 

The specific scheme is shown in \cref{fig:fig3}. Taking the traceable tag as an example, its corresponding lost information $L_{Trace}$ is also divided into two groups as input to the INN with the conditional feature from $I_{Tag}$. We employ the first two layers in Resnet18~\cite{60} as the feature extractor $\mathcal{O}_1(\cdot)$ and $\mathcal{O}_2(\cdot)$ to extract features as the conditions on the mapping of $L_{Trace}$ and $L_{Auth}$, respectively. For the $K$-th conditional invertible block of FAPM, the forward process of lost information $L_{Trace}$ is calculated as shown in \cref{eq:eq7} and \cref{eq:eq8}. The operation of authenticable tags is the same as the operation of traceable tags. The forward calculated process is shown in \cref{eq:eq9} and \cref{eq:eq10}.

\begin{equation}
    \vspace{-0.3cm}
	\begin{aligned}
		L_{\textit {Trace}_1}^{k+1}&=\exp \left(\alpha_1\left(L_{\textit {Trace}_2}^k, \mathcal{O}_1\left(I_{\textit {Tag }}\right)\right)\right) \odot L_{\textit {Trace}_1}^k \;\;\;\;\;\;\\
		&\;\;\;+\beta_1\left(L_{\textit {Trace}_2}^k, \mathcal{O}_1\left(I_{\textit {Tag }}\right)\right)
	\end{aligned}
	\label{eq:eq7}
% 	\vspace{-0.1cm}
\end{equation}

\begin{equation}
	\begin{aligned}
		L_{\textit {Trace}_2}^{k+1}=\exp \left(\gamma_1\left(L_{\textit {Trace}_1}^{k+1}\right)\right) \odot L_{\textit {Trace}_1}^k+\delta_1\left(L_{\textit {Trace}_1}^{k+1}\right)
	\end{aligned}
	\label{eq:eq8}
	\vspace{-0.2cm}
\end{equation}

\begin{equation}
	\begin{aligned}
		L_{\textit {Auth}_1}^{k+1}&=\exp \left(\alpha_2\left(L_{\textit {Auth}_2}^k, \mathcal{O}_2\left(I_{\textit {Tag }}\right)\right)\right) \odot L_{\textit {Auth}_1}^k\;\;\;\;\;\; \\
		&\;\;\;+\beta_2\left(L_{\textit {Auth}_2}^k, \mathcal{O}_2\left(I_{\textit {Tag }}\right)\right)
	\end{aligned}
	\label{eq:eq9}
	\vspace{-0.2cm}
\end{equation}

\begin{equation}
    % \vspace{cm}
	\begin{aligned}
		L_{\textit {Auth}_2}^{k+1}=\exp \left(\gamma_2\left(L_{\textit {Auth}_1}^{k+1}\right)\right) \odot L_{\textit {Auth}_1}^k+\delta_2\left(L_{\textit {Auth}_1}^{k+1}\right)
	\end{aligned}
	\label{eq:eq10}
\end{equation}

In the revealing process, the lost information is recovered through FAPM, and then the embedded tags are recovered from news images.

%-------------------------------------------------------------------------
\subsection{Distance Matric-Guided Module (DMGM)}

In the revealing process, we extract the traceable tag from all news images to verify the sources of news image. To effectively detect fake images, authenticable tag should be robust to moderate manipulation, while fragile to malicious manipulation, and the robustness can be evaluated by extraction error rates. Although extraction performance can be controlled by the designed loss function, the problem is more complex in a practical scenario since the types of manipulations are probably unseen unseen to the models. Thus, the generalization of the proposed model is still very important in fake news detection. Thus, we design a distance metric-guided modulation for tag extraction against the unseen types of manipulations.

The specific design is shown in \cref{fig:fig4}. We define the extracted correct authenticable tag from real news images as $T_{Auth}^{ext_{-}r}$ and the extracted wrong authenticable tag from fake news images as $T_{Auth}^{ext_{-}f}$. The target feature spaces are that $T_{Auth}^{ext_{-}r}$ should be near the origin and converge to a hypersphere of radius $r$ to maintain intra-class compactness. On the contrary, $T_{Auth}^{ext_{-}f}$ is expected to be away from the smaller hypersphere by a predefined margin $m$ to ensure inter-class separation between $T_{Auth}^{ext_{-}r}$ and $T_{Auth}^{ext_{-}f}$. The designed loss function is as follows, as inspired by the attack detection of unseen face presentation~\cite{59}.
\begin{equation}
\begin{aligned}
% 	\begin{gathered}
		\mathcal{L}_{\textit {margin }}&=\alpha_{\textit {Auth }}^{\textit {cor }} \times \max \left(d_{\text {Auth }}^{\textit {cor }}-r, 0\right) \\
		&\;\;\;+\alpha_{\textit {Auth }}^{\textit {wro }} \times \max \left((r+m)-d_{\textit {Auth }}^{\text {wro }}, 0\right)
% 	\end{gathered}
	\label{eq:eq11}
\end{aligned}
\end{equation}

Where, $d_{Auth}^{cor}=\left\|T_{Auth}^{ext_{-}r}-T_{Auth}\right\|_1^2$, $d_{Auth}^{wro}=\left\|T_{Auth}^{ext_{-}f}-T_{Auth}\right\|_1^2$. The distance is calculated between extracted tag samples and the embedded tags by using the square of $L$1-norm. Besides, $\alpha_{Auth}^{cor}$ and $\alpha_{Auth}^{wro}$ are the parameters to control the weights of loss contributed by extracted tag samples of different types.

%-------------------------------------------------------------------------
\subsection{Loss Function}

The total loss function is composed of four kinds of losses: the embedding and revealing losses of the main model, feature aware projection loss, distance metric-guided loss, and the low-frequency wavelet loss.

\noindent\textbf{Embedding loss}. The forward embedding process aims to embed $T_{Trace}$ and $T_{Auth}$ into original images $I_{Ori}$, to generate tagged news images $I_{Tag}$. To achieve this goal that $I_{Tag}$ should be indistinguishable from $I_{Ori}$, the embedding loss $L_{main}^{embed}$ is defined by
\begin{equation}
	\begin{gathered}
	\mathcal{L}_{\textit {main}}^{\textit {embed }}=\big\|I_{\textit {Ori}}-I_{Tag}\big\|
	\end{gathered}
	\label{eq:eq12}
\end{equation}
Where, $\mathcal{L}_{\textit {main }}^{\textit {embed }}$ measures the difference between original news images and tagged news images, and the difference can be measured $l_1$ or $l_2$ norm.

\noindent\textbf{Revealing loss}. The reveal process aims to extract the traceable tags from real news images and fake news images, and the authenticable tags from real news images. 
\begin{equation}
	\begin{gathered}
		\;\mathcal{L}_{Trace }^{ext\_r}=\Big\|T_{Trace}^{ext\_r}-T_{ Trace}\Big\|
	\end{gathered}
	\label{eq:eq13}
\end{equation}
\begin{equation}
	\begin{gathered}
		\;\mathcal{L}_{Trace }^{ext\_f}=\Big\|T_{Trace}^{ext\_f}-T_{ Trace}\Big\|
	\end{gathered}
	\label{eq:eq14}
\end{equation}
\begin{equation}
	\begin{gathered}
		\mathcal{L}_{Auth }^{ext\_r}=\Big\|T_{Auth}^{ext\_r}-T_{ Auth}\Big\|
	\end{gathered}
	\label{eq:eq15}
\end{equation}

Since we expect to extract the wrong authenticable tag from fake news images, there is no loss between $T_{Auth}^{ext_{-}f}$ and $T_{Auth}$.

\noindent\textbf{Feature aware projection loss}. We set the pre-defined distribution as uniform distribution $U(-1,1)$. 
\begin{equation}
	\begin{gathered}
		\mathcal{L}_{Z_{Trace}}=\mathcal{F}\left(Z_{\textit {Trace}}, y_{\text {z }}\right)
	\end{gathered}
	\label{eq:eq16}
\end{equation}
\begin{equation}
	\begin{gathered}
		\mathcal{L}_{Z_{Auth}}=\mathcal{F}\left(Z_{\textit {Auth}}, y_{\text {z }}\right)
	\end{gathered}
	\label{eq:eq17}
\end{equation}
Where, $\mathcal{F}$ is the pixel-level distance function, $y_{z}$ is the constant matrix, and $Z_{Trace}$ and $Z_{Auth}$ are the corresponding distributions.
%------------------------------------------------------------------------
\begin{table}[!t]
% 	\scriptsize           
	\centering
	\renewcommand\arraystretch{1.5}
	\tabcolsep=0.1cm
	
	\caption{
		The experimental results of different settings on parameters of $r$ and $m$.
	}\label{tab_3}
	\vspace{-0cm}
	\resizebox{\linewidth}{!}{
	\begin{tabular}{c|cccccc}
		%\begin{tabular}{|l|l|l|}
		%longtable
		\hline
		\multirow{2}{*}{\makecell[c]{Settings}} &  \multicolumn{2}{l}{\makecell[c]{News source tracing}} & 
		\multicolumn{2}{l}{\makecell[c]{Authenticity verification}} & 
		\multicolumn{2}{l}{\makecell[c]{Image quality}}\\
		\cline{2-7} & { $T_{Trace}^{ext\_r}$ } & { $T_{Trace}^{ext\_f}$  } & { $T_{Auth}^{ext\_r}$ } & { $T_{Auth}^{ext\_f}$  } & { PSNR } & { SSIM }\\ 
% 		\midrule
		\hline
		
		\shortstack{$r=0.1, m=0.2$}&  0.1200 & 0.1427 & 0.0700 & 0.0811 & 33.1790 & 0.8751 \\
		
		%\midrule
		
		\shortstack{$r=0.1, m=0.3$} & 0.0686 & 0.0939 & 0.0608 & 0.0722 & 34.9058 & 0.9074 \\
		
		%\midrule
		
		\shortstack{$r=0.07, m=0.15$} & \textbf{0.0275} & \textbf{0.0375} & \textbf{0.0558} & \textbf{0.2367} & \textbf{35.2095} & \textbf{0.9170} \\
		
		\hline
	\end{tabular}}
	\vspace{-0.3cm}
\end{table}

\begin{table*}[!t]
% 	\scriptsize       
    % \footnotesize
	\centering
	\tabcolsep=0.12cm
	\renewcommand\arraystretch{1.5}
	\caption{
		The extraction error rates of traceable tag and authenticable tag extracted from real/fake news images with different sizes, and the quality of tagged news images. 
	}
	\label{tab_1}
	\vspace{-0cm}
	\resizebox{0.95\textwidth}{!}{
% 	\begin{tabular}{c|cc|cc|cc|cc|cc|cc}
\begin{tabular}{c|cccccc|cccccc}
\hline
\multirow{3}{*}{Datasets} & \multicolumn{6}{c|}{128$\times$128}  & \multicolumn{6}{c}{256$\times$256}                                                                                                     \\ \cline{2-13} 
                          & \multicolumn{2}{c}{News source tracing} & \multicolumn{2}{c}{Authenticity verification} & \multicolumn{2}{c|}{Image quality} & \multicolumn{2}{c}{News source tracing} & \multicolumn{2}{c}{Authenticity verification} & \multicolumn{2}{c}{Image quality} \\ \cline{2-13} 
                          & { $T_{Trace}^{ext\_r}$ }\rule{0pt}{10pt} & { $T_{Trace}^{ext\_f}$  }\rule{0pt}{10pt} & { $T_{Auth}^{ext\_r}$ }\rule{0pt}{10pt} & { $T_{Auth}^{ext\_f}$  }\rule{0pt}{10pt} & { PSNR }\rule{0pt}{10pt} & { SSIM }\rule{0pt}{10pt}& { $T_{Trace}^{ext\_r}$ }\rule{0pt}{10pt} & { $T_{Trace}^{ext\_f}$  }\rule{0pt}{10pt} & { $T_{Auth}^{ext\_r}$ }\rule{0pt}{10pt} & { $T_{Auth}^{ext\_f}$  }\rule{0pt}{10pt} & { PSNR }\rule{0pt}{10pt} & { SSIM }\rule{0pt}{10pt}                 \\ \hline
                    %\midrule%\hline
        \shortstack{DIV2K~\cite{77}} & 0.0275 & 0.0375 & 0.0558 & 0.2367 & 35.2095 & 0.917 & 0.0017 & 0.0483 & 0.0542 & 0.2392 & 36.4556 & 0.9336 \\
		
		%\midrule
		
		\shortstack{COCO~\cite{78}} & 0.0407 & 0.0428 & 0.0929 & 0.2462 & 33.4367 & 0.8924 & 0.0262 & 0.0472 & 0.0977 & 0.3292 & 31.7688 & 0.8429 \\
		
		%\midrule
		
		\shortstack{Weibo~\cite{79}} & 0.0542 & 0.0649	 &	0.0918 & 0.1862 & 32.0736 &	0.8424 & 0.0298& 0.0572 & 0.1107  & 0.1937 &	31.8858 & 0.8056 \\
		
		%\midrule
		
		\shortstack{NewsStories~\cite{80}} & 0.0704 &	0.0600 & 0.1054 & 0.3373 & 34.0817 & 0.9176 &	0.0253  & 0.0535 & 0.0974  & 0.2334 & 31.4462 &	0.8160 \\
			
		%\midrule
		
		\shortstack{Flickr8k~\cite{81}} & 0.0658 & 0.0937&	0.1078  & 0.2985 & 33.5507 & 0.9051 &	0.0130 & 0.0142 & 0.0717  & 0.1970 & 32.0187&	0.7471 \\
		\hline
\end{tabular}}
\vspace{-0.3cm}
\end{table*}

\noindent\textbf{The low-frequency wavelet loss}. We employ the low-frequency wavelet loss $\mathcal{L}_{freq}$~\cite{54} to improve the quality and security of tagged news images. Suppose $\mathcal{H}(\cdot)_{LL}$ is the function of extracting low-frequency sub-bands after wavelet decomposition. The low-frequency wavlet loss is defined by
% \begin{equation}
% 	\begin{gathered}
% 		L_{Z_{Trace}}=F\left(Z_{\textit {Trace}}, y_{\text {z }}\right)
% 	\end{gathered}
% 	\label{eq:eq18}
% \end{equation}
\begin{equation}
	\begin{gathered}
		\mathcal{L}_{freq}(\theta)=\ell_{\mathcal{F}}\Big(\mathcal{H}(I_{Ori})_{LL}, \mathcal{H}(I_{Tag})_{LL}\Big)
	\end{gathered}
	\label{eq:eq18}
\end{equation}
Where, $\ell_{F}$ measures the difference between the low-frequency sub-bands of original news images and tagged news images.

\noindent\textbf{Total loss function}. The total loss function $\mathcal{L}_{total}$ is a weighted sum of losses in the main model, including the losses in feature aware projection, the hypersphere margin loss, and the low-frequency wavelet loss, which is represented by
% \begin{equation}
% 	\begin{gathered}
% 		L_{Z_{Trace}}=F\left(Z_{\textit {Trace}}, y_{\text {z }}\right)
% 	\end{gathered}
% 	\label{eq:eq19}
% \end{equation}

\begin{equation}
\begin{aligned}
% 	\begin{gathered}
		\mathcal{L}_{total} &= \omega_{1}\mathcal{L}^{embed}_{main} + \omega_{2}\mathcal{L}^{ext_{r}}_{Trace} + \omega_{3}\mathcal{L}^{ext_{f}}_{Trace} \\ & \;\;\;+ \omega_{4}\mathcal{L}^{ext_{r}}_{Auth} + \omega_{5}\mathcal{L}_{z_{Trace}} + \omega_{6}\mathcal{L}_{z_{Auth}} \\ &\;\;\;+\omega_{7}\mathcal{L}_{margin} + \omega_{8}\mathcal{L}_{freq}
% 	\end{gathered}
	\label{eq:eq19}
\end{aligned}
\end{equation}

% Where, $l_{F}$ measures the difference between the low-frequency sub-bands of original news images and tagged news images.

\section{Experimental Results and Analysis}
\subsection{Experimental Setups}
\noindent\textbf{Dataset}. The DIV2K training dataset~\cite{77} is used to train the proposed approach, which includes 800 high-resolution images. Before feeding the training images into the network, we first randomly crop the training images with the size of 128$\times$128. Next, we use the DIV2K test dataset to evaluate the performance, where the DIV2K test dataset consists of 100 high-resolution images. During the test process, we center-crop the test images with the size of 128$\times$128, 256$\times$256, and 1024$\times$1024 to verify our method can free different sizes. 

To further evaluate the generalization of the proposed approach, we use COCO 2017~\cite{78} validation dataset, and select 1,000 images from Weibo dataset~\cite{79}, NewsStories dataset~\cite{80}, and Flickr8k dataset ~\cite{81} separately to test the proposed method. 

\noindent\textbf{Setting the parameters}. In the network training, we use the Adam optimizer~\cite{59} with exponential decay rates of 0.9 and 0.999. Besides, we set the batch size to 8, the initial learning rate to 0.0002, and adjust the learning rate at 12,000 iterations with a decay penalty parameter of 0.1. The total training iterations are set to 15,000. In the training process, we design the moderate manipulations including jpeg compression $(Q=70)$, blur $(kernel=3)$, brightness $(b\sim U[-0.3,0.3])$, and contrast $(m\sim U[0.8,1.2])$. The malicious manipulations include splicing $(proportion\sim[0.36,0.49])$. Besides, we test different settings of parameters of $r$ and $m$ in DMGM with the image size 128$\times$128. As shown in \cref{tab_3}, when $r=0.07$ and $m=0.15$, the difference of $T_{Auth}^{ext_{-}r}$ and $T_{Auth}^{ext_{-}f}$ are distinguishable, and thus we set $r=0.07$ and $m=0.15$ in our experiments. Our experiments are performed by PyTorch 1.7 with an NVIDIA GeForce RTX 3090 GPU.

\subsection{News Images Authenticity Verification and Source Tracing Results}
To evaluate the performance of the proposed approach, we conduct the following experiments.

\noindent\textbf{News image authenticity verification}. To detect fake news images, \textit{i.e.}, maliciously manipulated news images, the extracted authenticable tags $T_{Auth}^{ext_{-}r}$ from real news images should be as similar as possible to its original one and $T_{Auth}^{ext_{-}f}$ should be far away from its original one. The higher extracted error rates of $T_{Auth}^{ext_{-}f}$ and the lower extracted error rates of $T_{Auth}^{ext_{-}r}$ mean better detection performance on fake news images. The experimental results are shown in \cref{tab_1}, it can be clearly observed that the extracted error rates of $T_{Auth}^{ext_{-}r}$ and $T_{Auth}^{ext_{-}f}$ have a large 
difference. The extracted error rates of  $T_{Auth}^{ext_{-}f}$ indicate that nearly half of the extracted bits are inaccurately extracted. The lower extraction error rates of  $T_{Auth}^{ext_{-}r}$ mean the embedded authenticable tag from real news images can be extracted accurately. The comparison of $T_{Auth}^{ext_{-}r}$ and $T_{Auth}^{ext_{-}f}$ show that the proposed approach can achieve desirable detection performance.

\begin{table}[!t]
	\scriptsize           
	\centering
	\tabcolsep=0.08cm
	\renewcommand\arraystretch{1.5}

	\caption{
		The experimental results for unseen manipulations. 
	}\label{tab_4}
	\vspace{-0cm}
	\begin{tabular}{cccc|cccc}
		%\begin{tabular}{|l|l|l|}
		\hline
		\multicolumn{4}{c|}{\makecell[c]{Unseen moderate manipulations}} 
		& \multicolumn{4}{c}{\makecell[c]{Unseen malicious manipulations}} \\
		\hline
		\multicolumn{2}{l}{\makecell[c]{Gaussian noise with \\ $\mu$=0, $\delta$=0.02}}%\rule{0pt}{10pt}
		& \multicolumn{2}{l|}{\makecell[c]{Saturation}}%\rule{0pt}{10pt} 
		& \multicolumn{2}{l}{\makecell[c]{Removal}}%\rule{0pt}{10pt} 
		& \multicolumn{2}{l}{\makecell[c]{Copy-move}}%\rule{0pt}{10pt} 
		\\ 
		\hline
		
		$T_{Trace}^{ext\_r}$ & $T_{Auth}^{ext\_r}$ & $T_{Trace}^{ext\_r}$ & $T_{Auth}^{ext\_r}$ & $T_{Trace}^{ext\_f}$ & $T_{Auth}^{ext\_f}$ & $T_{Trace}^{ext\_f}$ & $T_{Auth}^{ext\_f}$\\
		
		\hline
		
		0.0150 & 0.0480 & 0.0300 & 0.0450 & 0.0025 & 0.2558 & 0.0100 & 0.1850 \\
		
		\hline
	\end{tabular}
\end{table}

\begin{table}[!t]
% 	\scriptsize           
	\centering
	\tabcolsep=0.1cm
	\renewcommand\arraystretch{1.5}
	\caption{
		Performance comparison with SoTA methods. 
	}\label{tab_5}
	\vspace{-0cm}
	\resizebox{\linewidth}{!}{
	\begin{tabular}{c|cccccc}
		%\begin{tabular}{|l|l|l|}
		%longtable
% 		\toprule
		\hline
		\multirow{2}{*}{\makecell[c]{Method}} &  \multicolumn{2}{l}{\makecell[l]{News source tracing}} & 
		\multicolumn{2}{l}{\makecell[c]{Authenticity verification}} & 
		\multicolumn{2}{l}{\makecell[c]{Image quality}}\\
		\cline{2-7} & { $T_{Trace}^{ext\_r}$ }\rule{0pt}{10pt} & { $T_{Trace}^{ext\_f}$  }\rule{0pt}{10pt} & { $T_{Auth}^{ext\_r}$ }\rule{0pt}{10pt} & { $T_{Auth}^{ext\_f}$  }\rule{0pt}{10pt} & { PSNR }\rule{0pt}{10pt} & { SSIM }\rule{0pt}{10pt}\\ 
% 		\midrule
        \hline
		
		\shortstack{ISN~\cite{55}} & 0.2651 & 0.5229 & 0.5013 & 0.4537 & 34.6334 & 0.9747 \\
		
		%\midrule
		
		\shortstack{HiNet~\cite{54}} & 0.2158 & 0.1725 & 0.5333 & 0.5409 & 29.0537 & 0.7984 \\
		
		%\midrule
		
		\makecell{MBRS~\cite{70}\\ w/ preprocess} & 0.3366 & 0.5333 & 0.4183 & 0.5333 & 35.2361 & 0.9764 \\
		
		%\midrule
		
		\makecell{MBRS~\cite{70}\\ w/o preprocess} & 0.0375 & 0.0458 & 0.0583 & 0.0608 & 36.2282 & 0.9817 \\
			
		%\midrule
		
		\shortstack{TRLH~\cite{82}} & 0.1983 & 0.1864 & 0.1836 & 0.1751 & \textbf{46.3329} & \textbf{0.9903} \\
			
		%\midrule
		
		\shortstack{Ours} & \textbf{0.0017} & \textbf{0.0483} & \textbf{0.0542} & \textbf{0.2392} & 36.4556 & 0.9336 \\
		
% 		\bottomrule
		\hline
	\end{tabular}
	}
	\vspace{-0cm}
\end{table}
\begin{table*}[!t]
	\scriptsize           
	\centering
	\renewcommand\arraystretch{1.2}
	\caption{
		Ablation studies for every model design, including DINN, FAPM, and DMGM. The performances of news authenticity verification and source tracing are evaluated by the error rates of traceable tags and authenticable tags extracted from news images. The tagged news image quality is evaluated by PSNR and SSIM metrics.
	}\label{tab_2}
	\vspace{-0cm}
    
	    \begin{tabular}{c|ccc|cccccc}
		%\begin{tabular}{|l|l|l|}
		%longtable
		\hline
		\multirow{2}{*}[-0.08cm]{\makecell[c]{Size of news \\ images}} & \multirow{2}{*}[-0.08cm]{DINN} & \multirow{2}{*}[-0.08cm]{FAPM} & \multirow{2}{*}[-0.08cm]{DMGM} & \multicolumn{2}{l}{\makecell[c]{News source tracing}} & 
		\multicolumn{2}{l}{\makecell[c]{Authenticity verification}} & 
		\multicolumn{2}{l}{\makecell[c]{Image quality}}\\
		\cline{5-10} &  &  &  &{ $T_{Trace}^{ext\_r}$ }\rule{0pt}{10pt} & { $T_{Trace}^{ext\_f}$  }\rule{0pt}{10pt} & { $T_{Auth}^{ext\_r}$ }\rule{0pt}{10pt} & { $T_{Auth}^{ext\_f}$  }\rule{0pt}{10pt} & { PSNR }\rule{0pt}{10pt} & { SSIM }\rule{0pt}{10pt}\\ 
		\hline%\midrule
		
		\multirow{4}{*}[-0.15cm]{\makecell[c]{$128 \times 128$}} 
		& {$\times$}\rule{0pt}{10pt} & $\checkmark$ & $\checkmark$ & 0.4868 & 0.5138 & 0.2942 & 0.3083 & 19.9774 & 0.4711 \\
		
        & {$\checkmark$}\rule{0pt}{10pt} & $\times$ & $\checkmark$ & 0.0741 & 0.1043 & 0.0693 & 0.0938 & \textbf{37.4220} & \textbf{0.9430}\\
		
% 		{\shortstack{128$\times$128}} & {$\times$}\rule{0pt}{10pt} & $\checkmark$ & $\checkmark$ & 0.4868 & 0.5138 & 0.2942 & 0.3083 & 19.9774 & 0.4711 \\
		
% 		& {$\checkmark$}\rule{0pt}{10pt} & $\times$ & $\checkmark$ & 0.0741 & 0.1043 & 0.0693 & 0.0938 & \textbf{37.4220} & \textbf{0.9430}\\
		
    	& {$\checkmark$}\rule{0pt}{10pt} & $\checkmark$ & $\times$ & 0.0650 & 0.0870 & 0.0636 & 0.0327 & 36.0256 & 0.9223\\
    	
    	& {$\checkmark$}\rule{0pt}{10pt} & $\checkmark$ & $\checkmark$ & \textbf{0.0275} & \textbf{0.0375} & \textbf{0.0558} & \textbf{0.2367} & 35.2095 & 0.9170\\
    	
		\hline
		
		\multirow{4}{*}[-0.15cm]{\makecell[c]{$256 \times 256$}} & {$\times$}\rule{0pt}{10pt} & $\checkmark$ & $\checkmark$ & 0.5043 & 0.5297 & 0.2497 & 0.2647 & 19.4587 & 0.4178 \\
		
		& {$\checkmark$}\rule{0pt}{10pt} & $\times$ & $\checkmark$ & 0.0647 & 0.0997 & 0.0668 & 0.0800 & 36.3945 & 0.9350\\
		
		& {$\checkmark$}\rule{0pt}{10pt} & $\checkmark$ & $\times$ & 0.0317 & 0.0475 & 0.0517 & 0.0833 & \textbf{39.1990} & \textbf{0.9620}\\
		
		& {$\checkmark$}\rule{0pt}{10pt} & $\checkmark$ & $\checkmark$ & \textbf{0.0017} & \textbf{0.0483} & \textbf{0.0542} & \textbf{0.2392} & 36.4556 & 0.9336\\
		
		\hline
		
    	\multirow{4}{*}[-0.15cm]{\makecell[c]{$1024 \times 1024$}} & {$\times$}\rule{0pt}{10pt} & $\checkmark$ & $\checkmark$ & 0.4245 & 0.4380 & 0.4413 & 0.4390 & 20.2916 & 0.3883 \\
	
    	& {$\checkmark$}\rule{0pt}{10pt} & $\times$ & $\checkmark$ & 0.0402 & 0.0480 & 0.0680 & 0.0625 & 38.6636 & 0.9507\\
	
	    & {$\checkmark$}\rule{0pt}{10pt} & $\checkmark$ & $\times$ & 0.1367 & 0.1567 & 0.0317 & 0.0375 & \textbf{39.7316} & \textbf{0.9606}\\
	
    	& {$\checkmark$}\rule{0pt}{10pt} & $\checkmark$ & $\checkmark$ & \textbf{0.0192} & \textbf{0.0141} & \textbf{0.0342} & \textbf{0.1883} & 33.1989 & 0.8691\\
		\hline
	    \end{tabular}
	    \vspace{-0.3cm}
\end{table*}

\noindent\textbf{News source traceability}. To trace news sources, the traceable tag should be robust to all manipulations. Therefore, lower extracted error rate means better traceability performance. As shown in \cref{tab_1}, the extraction error rate of $T_{Trace}^{ext_{-}f}$ is slightly higher than that of $T_{Trace}^{ext_{-}r}$, since the fake news images undergo malicious manipulations. However, it can be observed that all extraction error rates of $T_{Trace}^{ext_{-}r}$ and $T_{Trace}^{ext_{-}f}$ remain below 0.01. The extraction performance of traceable tags shows the effectiveness of news traceability. 

\noindent\textbf{Visual results of tagged news images}. For news images, the quality of tagged images needs to be maintained. We conduct experiments to test the quality of tagged news images by using PSNR and SSIM. The results are shown in \cref{tab_1}. For the different sizes of news images, the values of PSNR on the tagged news images are about 35 and that of SSIM are more than 0.8, which indicate the quality of tagged images maintain at a high-level.

\noindent\textbf{The generalization ability on diverse datasets}. We test the generalization ability of the proposed tagging approach on different datasets. As shown in \cref{tab_1}, the extraction performance of traceable and authenticable tags is consistent with the above analysis, which can achieve news image authenticity verification and source tracing on diverse datasets.

\noindent\textbf{The generalization ability on anti-unseen manipulations}. We test the generalization ability of proposed tagging approach on anti-unseen manipulations. We test dual-tags on the unseen moderate manipulations, \textit{i.e.}, Gaussian noise with ${\mu}=0,
{\delta}=0.02$, and saturation with randomly linearly interpolating between the full RGB image and its grayscale equivalent. Also, we test the dual-tags by adding the removal and copy-move as unseen malicious manipulations. The experimental results are shown in \cref{tab_4}. It can be observed that there is a slight increase in the extraction error rates, the extraction error rates of $T_{Trace}^{ext\_r}$, $T_{Trace}^{ext\_f}$, and $T_{Auth}^{ext\_r}$ are still much lower than that of $T_{Auth}^{ext\_f}$. The extraction performance of $T_{Trace}^{ext\_r}$, $T_{Trace}^{ext\_f}$, and $T_{Auth}^{ext\_r}$ is desirable. The above observations indicate that the proposed approach has good generalizability on popular unseen manipulations.

\noindent\textbf{Comparison with SoTA approaches}. We compare our approach with the State-of-The-Art (SoTA) approaches, including two INN-based hiding algorithms, \textit{i.e.}, ISN~\cite{55} and HiNet~\cite{54}, an autoencoder-based hiding approach, \textit{i.e.}, MBRS~\cite{70}, and a dual watermarking approach, \textit{i.e.}, TRLH~\cite{82}. Note that ISN~\cite{55} and HiNet~\cite{54} are originally designed to embed images, we preprocess the bit messages to decimal numbers to narrow the distribution gap between the embedded data with that of images. As MBRS~\cite{70} is originally designed to embed binary bit messages, we test the performance of both embedding of the preprocessed decimal numbers and the binary bits. For TRLH~\cite{82}, we keep its original settings.

The experimental results are shown in \cref{tab_5}. It can be observed that the extraction performances of traceable tags and authenticable tags of proposed approach outperforms those of SoTA approaches under both moderate and malicious manipulations. Note that only the proposed approach shows distinguishable differences between the extraction of traceable tags and authenticable tags under malicious attacks. In summary, our approach achieves the SoTA performances for both news traceability and authenticity verification in fake news image detection. Note that the image quality of HiNet is inferior, because the solid color image-like hidden information is not considered in their training.

\subsection{Ablation Study}
\noindent\textbf{Effect of DINN}. When the classical INN in HiNet \cite{54} is used as the baseline to verify the effect of DINN, as shown in \cref{tab_2}, the extraction error rates of all embedded tags are very high regardless of the sizes of the news images, especially for the traceable tag, which are above 0.4. The high extraction error rates mean that it is hard to verify the news source. Besides, such close extraction error rates of $T_{Auth}^{ext\_r}$ and $T_{Auth}^{ext\_f}$ make it hard to detect fake news images. According to this phenomenon, if the network uses the same set of parameters to embed the dual-tags, it can cause conflicts and contradictions with network parameters.

By comparing the experimental results of the proposed DINN, there are low extraction error rates for $T_{Trace}^{ext\_r}$, $T_{Trace}^{ext\_f}$, and $T_{Auth}^{ext\_r}$ while high extraction error rates for $T_{Auth}^{ext\_f}$. That indicates the proposed approach can achieve fake news image detection and source tracing. Besides, the image quality of the tagged news images is also improved. 

\noindent\textbf{Effect of FAPM}. The experimental results for the comparison with/without FAPM are shown in \cref{tab_2}. Without FAPM, tagged news images have a better performance on image quality and a decrease in the extraction error rates for all tags. However, the extract error rates of $T_{Auth}^{ext\_f}$ are also very low, which make it hard to distinguish it from $T_{Auth}^{ext\_r}$. So, fake news images cannot be detected. By comparing the proposed approach with FAPM, the extraction error rates of $T_{Auth}^{ext\_f}$ have a much higher value than those of $T_{Auth}^{ext\_r}$. Thus, it can be used to verify the news authenticity.

\noindent\textbf{Effect of DMGM}. To increase the difference between the extraction error rates of $T_{Auth}^{ext\_r}$ and $T_{Auth}^{ext\_f}$, we design a margin loss to control the range of extraction error rates. From \cref{tab_2}, when there is no margin loss for the distance metric to increase the difference between $T_{Auth}^{ext\_r}$ and $T_{Auth}^{ext\_f}$, the extraction error rates of them are very close whereas the proposed DMGM creates a large margin on the extraction error rates of $T_{Auth}^{ext\_r}$ and $T_{Auth}^{ext\_f}$. It makes fake news image detection more accurate. 

% \vspace{-0.3cm}
\section{Conclusion and Discussion}
In this paper, we have presented a novel proactive invisible tagging approach for reliable fake news detection. The main purpose is to extract pre-embedded dual-tags, \textit{i.e.}, traceable tags and authenticable tags from news images to verify the news authenticity and trace the news source for reliable fake news detection. To achieve the above goals simultaneously, the DINN is designed to decouple the embedding and extracting processes of the dual-tags with no contradiction in network parameters. Besides, the designed FAPM and DMGM provide the dual-tags with different robustness performances and improve the tag extraction performance. In the future, we will integrate news textual information with the proposed tagging approach for fake news detection and enhance the extraction performance of tags under unseen manipulations.

\iffalse
\begin{table*}[!t]
	\scriptsize           
	\centering
	\caption{
	    The extraction error rates from real/fake news images totrace the news source and detect the fake news. And the image quality performance of tagged news images.
	}\label{tab_1}
	
	\begin{tabular}{ccccccc}
		%\begin{tabular}{|l|l|l|}
		%longtable
		\toprule%\hline
		\multirow{2}{*}{\makecell[c]{Size of news \\ images}} &  \multicolumn{2}{l}{\makecell[c]{Extraction error rates \\ from real news images}} & 
		\multicolumn{2}{l}{\makecell[c]{Extraction error rates \\ from fake news images}} & 
		\multicolumn{2}{l}{\makecell[c]{Quality of Tagged News \\ Images}}\\
		\cline{2-7} & { $T_{Trace}^{ext\_r}$ }\rule{0pt}{10pt} & { $T_{Auth}^{ext\_r}$ }\rule{0pt}{10pt} & { $T_{Auth}^{ext\_f}$ }\rule{0pt}{10pt} & { $T_{Trace}^{ext\_r}$ }\rule{0pt}{10pt} & { $T_{Auth}^{ext\_r}$ }\rule{0pt}{10pt} & { $T_{Trace}^{ext\_f}$ }\rule{0pt}{10pt}\\ 
		\midrule%\hline
		\shortstack{128$\times$128} & 0.0558 & 0.0558 & 0.0375 & 0.2367 & 35.2095 & 0.9170 \\
		
		\midrule
		
		\shortstack{256$\times$256}&  0.0017 & 0.0542 & 0.0483 & 0.2392 & 36.4556 & 0.9336 \\
		
		\midrule
		
		\shortstack{1024$\times$1024} & 0.0192 & 0.0342 & 0.0141 & 0.1883 & 34.1989 & 0.9091 \\
		
		\bottomrule%\hline
	\end{tabular}
\end{table*}
\fi

%%%%%%%%% REFERENCES
{\small
\bibliographystyle{ieee_fullname}
\bibliography{Paper}
}

\end{document}